\documentclass{article}

\makeatletter

\usepackage[
  top=2.25cm, bottom=2.5cm,
  left=2.5cm,  right=2.5cm
]{geometry}
\usepackage{microtype}
\usepackage{placeins}
\usepackage{hyphenat}
\usepackage{setspace}
\usepackage{etoolbox}

\usepackage{ifxetex}
\usepackage{ifluatex}

\ifxetex
  \usepackage{fontspec}
  \defaultfontfeatures{Ligatures=TeX}
  \IfFontExistsTF{Palatino}{%
    \setmainfont{Palatino}
    \setsansfont{Helvetica Neue}[Scale=MatchLowercase]
    \setmonofont{Menlo}[Scale=MatchLowercase]
  }{}
  \usepackage{xeCJK}
  \xeCJKsetup{CJKmath=true}
  \IfFontExistsTF{PingFang SC}{%
    \setCJKmainfont[BoldFont=PingFang SC Semibold, ItalicFont=Kaiti SC]{PingFang SC}
    \setCJKsansfont{PingFang SC}
    \setCJKmonofont{PingFang SC}
  }{%
    \IfFontExistsTF{Noto Serif CJK SC}{%
      \setCJKmainfont{Noto Serif CJK SC}
      \setCJKsansfont{Noto Sans CJK SC}
      \setCJKmonofont{Noto Sans Mono CJK SC}
    }{%
      \IfFontExistsTF{SimSun}{%
        \setCJKmainfont{SimSun}
        \setCJKsansfont{SimHei}
        \setCJKmonofont{FangSong}
      }{%
        \PackageWarning{arxiv}{No CJK font found; Chinese text may not render}%
        \setCJKmainfont{Noto Serif CJK SC}%
      }%
    }%
  }
\else\ifluatex
  \usepackage{fontspec}
  \defaultfontfeatures{Ligatures=TeX}
  \usepackage{luatexja-fontspec}
\else
  \usepackage[T1]{fontenc}
  \usepackage[utf8]{inputenc}
  \usepackage{mathpazo}
  \usepackage{helvet}
  \usepackage{courier}
  \usepackage{CJKutf8}
\fi\fi

\ifxetex
  \newenvironment{CJKblock}{}{}
\else\ifluatex
  \newenvironment{CJKblock}{}{}
\else
  
\fi\fi

\usepackage{graphicx}
\usepackage{subcaption}
\usepackage{tikz}
\usetikzlibrary{arrows.meta, positioning, shapes, shapes.geometric, calc, backgrounds, shadows, fit}

\usepackage{booktabs}
\usepackage{array}
\usepackage{tabularx}
\usepackage{longtable}
\usepackage{multirow}
\usepackage{enumitem}
\setlist[itemize]{leftmargin=1.3em,itemsep=0.2em,topsep=0.2em}
\setlist[enumerate]{leftmargin=1.5em,itemsep=0.2em,topsep=0.2em}

\usepackage{amsmath}
\usepackage{amssymb}
\usepackage{amsfonts}
\usepackage{bm}

\usepackage{xurl}

\usepackage{xcolor}
\usepackage[most]{tcolorbox}
\definecolor{reportblue}{HTML}{0091FF}
\definecolor{reportink}{HTML}{1C2B33}
\definecolor{reportmuted}{HTML}{5F6F7A}
\definecolor{reportbg}{HTML}{EAF4FB}
\definecolor{reportline}{HTML}{B9DFF5}
\definecolor{reportorange}{HTML}{FF6A00}
\definecolor{reportgreen}{HTML}{00A86B}
\definecolor{reportred}{HTML}{E8453C}

\usepackage[round,authoryear]{natbib}

\usepackage{hyperref}
\hypersetup{
  colorlinks,
  linkcolor  = reportblue,
  citecolor  = reportblue,
  urlcolor   = reportblue
}
\usepackage[noabbrev,nameinlink]{cleveref}

\usepackage{titlesec}
\titleformat{\section}
  {\Large\sffamily\bfseries\color{reportink}}
  {\thesection}{0.8em}{}
\titleformat{\subsection}
  {\large\sffamily\bfseries\color{reportink}}
  {\thesubsection}{0.8em}{}
\titleformat{\subsubsection}
  {\normalsize\sffamily\bfseries\color{reportink}}
  {\thesubsubsection}{0.8em}{}
\titleformat*{\paragraph}{\sffamily\bfseries}
\titlespacing*{\section}{0pt}{1.1em}{0.55em}
\titlespacing*{\subsection}{0pt}{0.9em}{0.35em}
\titlespacing*{\subsubsection}{0pt}{0.7em}{0.25em}

\usepackage{caption}
\DeclareCaptionLabelSeparator{custom}{}
\DeclareCaptionFormat{custom}{{\sffamily\textbf{#1 #2}} #3}
\usepackage{fancyhdr}
\fancypagestyle{plain}{%
  \fancyhf{}
  
  \fancyfoot[C]{\small\color{reportmuted}\thepage}
}

\fancypagestyle{firstpage}{%
  \fancyhf{}
  
  \fancyfoot[C]{}
}

\newcommand\report@addtolist[5][]{%
  \begingroup
    \if\relax#3\relax\def\sep{}\else\def\sep{#5}\fi
    \let\protect\@unexpandable@protect
    \xdef#3{\expandafter{#3}\sep #4[#1]{#2}}%
  \endgroup
}

\newcommand\authorlist{}
\newcommand\authorformat[2][]{{\sffamily\bfseries #2$^{#1}$}}
\renewcommand\author[2][]{\report@addtolist[#1]{#2}{\authorlist}{\authorformat}{, }}

\newcommand\affiliationlist{}
\newcommand\affiliationformat[2][]{{\normalsize $^{#1}$\,#2}}
\newcommand\affiliation[2][]{\report@addtolist[#1]{#2}{\affiliationlist}{\affiliationformat}{, }}

\newcommand\contributionlist{}
\newcommand\contributionformat[2][]{{\small\color{reportmuted} $^{#1}$#2}}
\newcommand\contribution[2][]{\report@addtolist[#1]{#2}{\contributionlist}{\contributionformat}{, }}

\newcommand\metadatalist{}
\newcommand\metadataformat[2][]{{\small{\sffamily\bfseries #1:} #2}}
\newcommand\metadata[2][]{\report@addtolist[#1]{#2}{\metadatalist}{\metadataformat}{\par}}

\newcommand{\abstractlist}{}
\renewcommand{\abstract}[1]{\gdef\abstractlist{{\color{reportink} #1}}}
\newcommand{\email}[1]{\href{mailto:#1}{\texttt{#1}}}
\renewcommand\date[1]{\metadata[Date]{#1}}

\newcommand{\titlelist}{{\huge\sffamily\bfseries Untitled Technical Report}}
\renewcommand{\title}[1]{\gdef\titlelist{{\huge\sffamily\bfseries #1}}}

\newcommand{\reportlogofile}{yuan-title-logo.png}

\newlength{\reportlogosize}
\newcommand{\mymaketitle}{%
  \thispagestyle{firstpage}%
  \tcbset{enhanced,frame hidden}
  \tcbset{left=0.5cm, right=0.5cm, top=0.5cm, bottom=0.5cm}
  \tcbset{arc=10pt, colback=reportbg}
  \tcbset{before skip=0pt}
  \tcbset{grow to left by=1.5pt, grow to right by=1.5pt}
  \tcbset{overlay={\node[
    anchor=north east,
    at=(frame.north east),
    xshift=-0.18cm,
    yshift=-0.18cm,
    inner sep=0pt] {%
      \begin{tikzpicture}
        \clip[rounded corners=0.35cm]
          (0,0) rectangle (\reportlogosize,\reportlogosize);
        \node[anchor=center,inner sep=0pt]
          at (0.5\reportlogosize,0.5\reportlogosize)
          {\includegraphics[
            width=\dimexpr\reportlogosize-0.15cm\relax,
            height=\dimexpr\reportlogosize-0.15cm\relax,
            keepaspectratio]{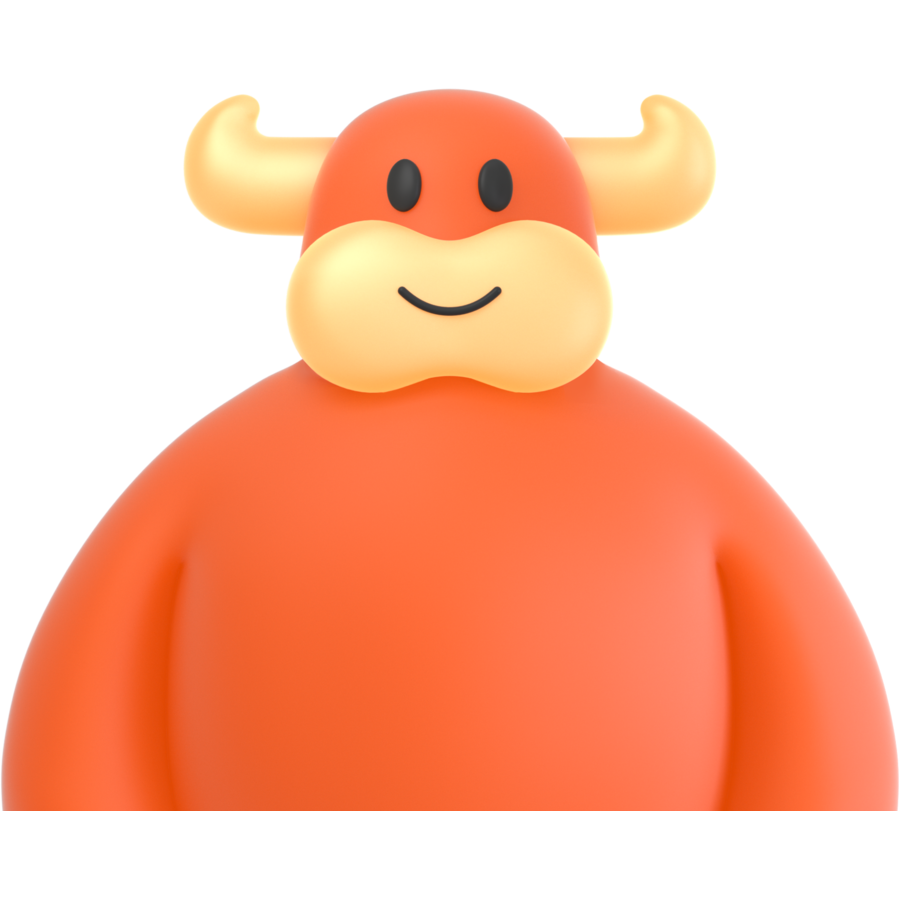}};
      \end{tikzpicture}%
    };}}%
  \begin{tcolorbox}
    \setlength{\parindent}{0cm}
    \setlength{\parskip}{0.5cm}
    {%
      \setlength{\parskip}{0cm}
      \raggedright
      \nohyphens
      {%
        \setstretch{1.5}
        \begin{minipage}{\dimexpr\linewidth-\reportlogosize-0.3cm\relax}
          \raggedright
          \titlelist
        \end{minipage}\par
      }%
      \vskip 0.2cm
      \authorlist\par
      \vskip 0.2cm
      \affiliationlist\par
      \ifdefempty{\contributionlist}{}{\contributionlist\par}
    }%
    \abstractlist\par
    \vskip 0.5cm
    {%
      \setlength{\parskip}{0cm}
      \ifdefempty{\metadatalist}{\vspace*{0.65cm}}{\metadatalist\par}
    }%
  \end{tcolorbox}
  \tcbset{reset}
  \FloatBarrier
}

\if@twocolumn
\renewcommand{\maketitle}{%
  \twocolumn[%
    \mymaketitle
    \vskip 0.38cm
  ]%
}
\else
\renewcommand{\maketitle}{%
  \mymaketitle
}
\fi

\makeatother

\usepackage{colortbl}
\usepackage{float}

\newcommand{\bench}{IndustryBench-\textsc{MIPU}}
\definecolor{bestrow}{RGB}{238,238,255}
\newcommand{\tablefont}{\small}
\newcommand{\tighttable}{\tablefont\renewcommand{\arraystretch}{1.08}}
\newcolumntype{Y}{>{\raggedright\arraybackslash}X}
\newcolumntype{Z}{>{\raggedleft\arraybackslash}X}

\title{\bench{}: Benchmarking Multi-Image Attribute Value Extraction for Industrial Products}

\author[]{Multimodal and Industrial AI Team\hyperref[sec:authors]{\textsuperscript{\textdagger}}}
\affiliation[]{Taobao\&Tmall, Alibaba Group}

\date{June 2026}
\metadata[Paper]{\href{https://arxiv.org/abs/2606.14383}{arxiv.org/abs/2606.14383}}
\metadata[Dataset]{\href{https://huggingface.co/datasets/alibaba-multimodal-industrial-ai/IndustryBench-MIPU}{huggingface.co/datasets/alibaba-multimodal-industrial-ai/IndustryBench-MIPU}}
\metadata[Code]{\href{https://github.com/alibaba-multimodal-industrial-ai/IndustryBench-MIPU}{github.com/alibaba-multimodal-industrial-ai/IndustryBench-MIPU}}

\abstract{%
Industrial products such as valves and circuit breakers are defined by dense technical specifications that govern procurement, compatibility, and safety across supply chains. These specifications are scattered across multiple heterogeneous product images, including specification tables, nameplates, and technical drawings, yet whether Multimodal Large Language Models (MLLMs) can reliably recover them remains underexplored. To fill this gap, we introduce \bench{}, the first large-scale benchmark for multi-image industrial product understanding, built around structured attribute extraction---recovering property-value pairs from product images. This task jointly probes text recognition on specification tables and nameplates, visual reasoning over technical drawings, domain knowledge to decode industrial terminology, and cross-image evidence integration to assemble scattered specifications. Concretely, the benchmark comprises 4{,}559 products across 27{,}652 images with 103{,}703 annotations spanning 18 industrial categories, constructed through multi-model consensus and three-tier quality assurance. Evaluating nine MLLMs under both single-image and product-level multi-image settings reveals a stark \textit{completeness gap}: models achieve high precision (\textbf{86--94\%}) but the best recovers only \textbf{49.9\%} of product-level attributes; moving from single-image to multi-image extraction costs \textbf{15--34} percentage points of recall. Multi-image completeness, not single-image accuracy, is the core bottleneck. Dataset and code are publicly available.%
}

\begin{document}
\setlength{\emergencystretch}{3em}
\widowpenalty=10000
\clubpenalty=10000
\displaywidowpenalty=10000

\maketitle

% ============================================================
% 1. INTRODUCTION
% ============================================================
\section{Introduction}
\label{sec:intro}

Industrial products such as valves, circuit breakers, chemical reagents, and steel alloys are defined by dense technical specifications that govern purchasing decisions, product compatibility, and operational safety across manufacturing and supply chains. As Multimodal Large Language Models (MLLMs) are increasingly deployed to automate procurement, inventory cataloging, and cross-supplier product matching \citep{balkan2025procurement,chen2024bimcompliance}, a critical question emerges: \textit{can these models understand the complex, knowledge-intensive specifications of industrial products distributed across multiple product images?}

\begin{figure}[t]
\centering
\includegraphics[width=\textwidth]{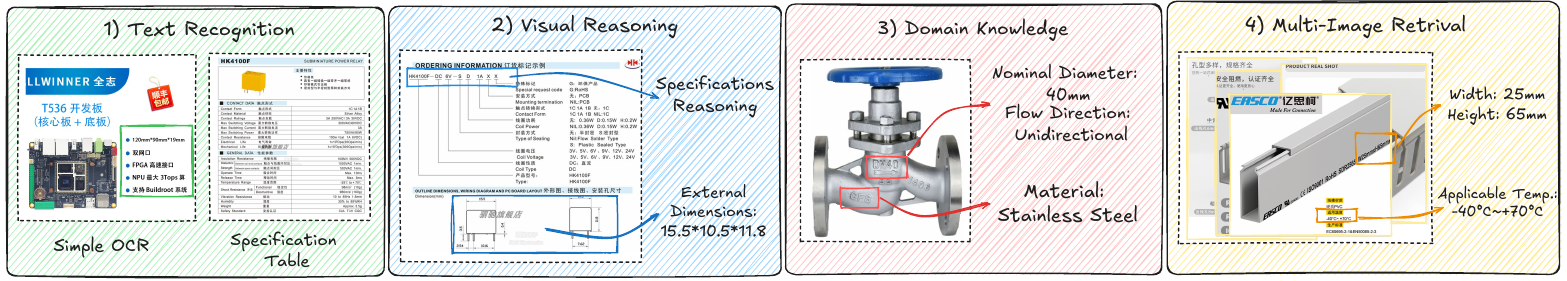}
\caption{Four Challenges in Multi-Image Industrial Product Understanding: (1)~text recognition from labels and specification tables; (2)~visual reasoning over technical drawings; (3)~domain knowledge interpretation; (4)~cross-image evidence integration.}
\label{fig:intro}
\end{figure}

Figure~\ref{fig:intro} illustrates why this question is challenging. Industrial product specifications appear in dense tables, nameplates, and labels, requiring reliable text recognition. Beyond surface-level reading, technical drawings demand visual reasoning to associate measurements, arrows, and structural annotations with product attributes. Many attributes further require domain knowledge interpretation: ``DN20'' denotes nominal diameter and ``304'' indicates a steel grade. These challenges compound in the multi-image setting: relevant evidence is scattered across multiple images interspersed with irrelevant content, forcing models to locate and integrate specification fragments across heterogeneous visual sources.

Despite these challenges being central to real-world industrial scenarios, existing benchmarks address only fragments of the problem. General-purpose MLLM benchmarks evaluate broad visual reasoning \citep{yue2024mmmu,liu2023mmbench,meng2025mmiu} or text recognition \citep{liu2024ocrbench,singh2019textvqa,mathew2021docvqa}, but not the domain knowledge required to interpret technical specifications. Attribute value extraction benchmarks target text, single product images, or consumer goods \citep{zheng2018opentag,xu2019scaling,yang2022mave,zhu2020multimodal,wang2024implicitave}, but not multi-image industrial profiles. Industrial AI benchmarks focus on defect detection and assembly recognition \citep{bergmann2019mvtec,forge2026} or text-only domain QA \citep{industrybench2026}, without evaluating multimodal specification comprehension. No prior work jointly evaluates text recognition, visual reasoning, domain knowledge, and cross-image evidence integration in the industrial product domain.

To fill this gap, we introduce \bench{} (Multi-Image Industrial Product Understanding Benchmark), the first multi-image benchmark for evaluating MLLM understanding of industrial products. We adopt structured attribute extraction---recovering property-value pairs from product images given a product-specific schema---as the evaluation task, since it naturally engages all four challenges. Concretely, extracting a complete product record requires reading specification tables, interpreting technical drawings, decoding domain terminology, and locating evidence across images, while the structured output remains schema-constrained and objectively scorable. We construct the benchmark through a semi-automated pipeline with multi-model consensus and three-tier quality assurance. Evaluation proceeds under both single-image and product-level multi-image settings to separately probe localized extraction and cross-image integration. Evaluating nine MLLMs reveals a substantial \textit{completeness gap}: no model exceeds \textbf{50\%} recall in the multi-image setting, and moving from single-image to multi-image extraction costs \textbf{15--34} percentage points of recall, demonstrating that cross-image evidence integration---not per-image recognition---is the primary bottleneck.

Our contributions are as follows:
\begin{itemize}[nosep,leftmargin=*]
  \item \textbf{Task formalization.} We formalize multi-image industrial product attribute extraction and define two evaluation settings---single-image and product-level multi-image---that separately probe localized extraction and cross-image integration (\S\ref{sec:task}).
  \item \textbf{Benchmark.} We construct \bench{}, a large-scale benchmark spanning 18 industrial categories with over 100K product-level annotations, built through multi-model consensus and three-tier quality assurance (\S\ref{sec:benchmark}).
  \item \textbf{Findings.} We benchmark nine MLLMs under both settings, showing that current models are precise but substantially incomplete, and that the completeness bottleneck is driven by cross-image evidence integration rather than per-image recognition (\S\ref{sec:experiments}).
\end{itemize}

\begin{figure}[t]
\centering
\includegraphics[width=\textwidth]{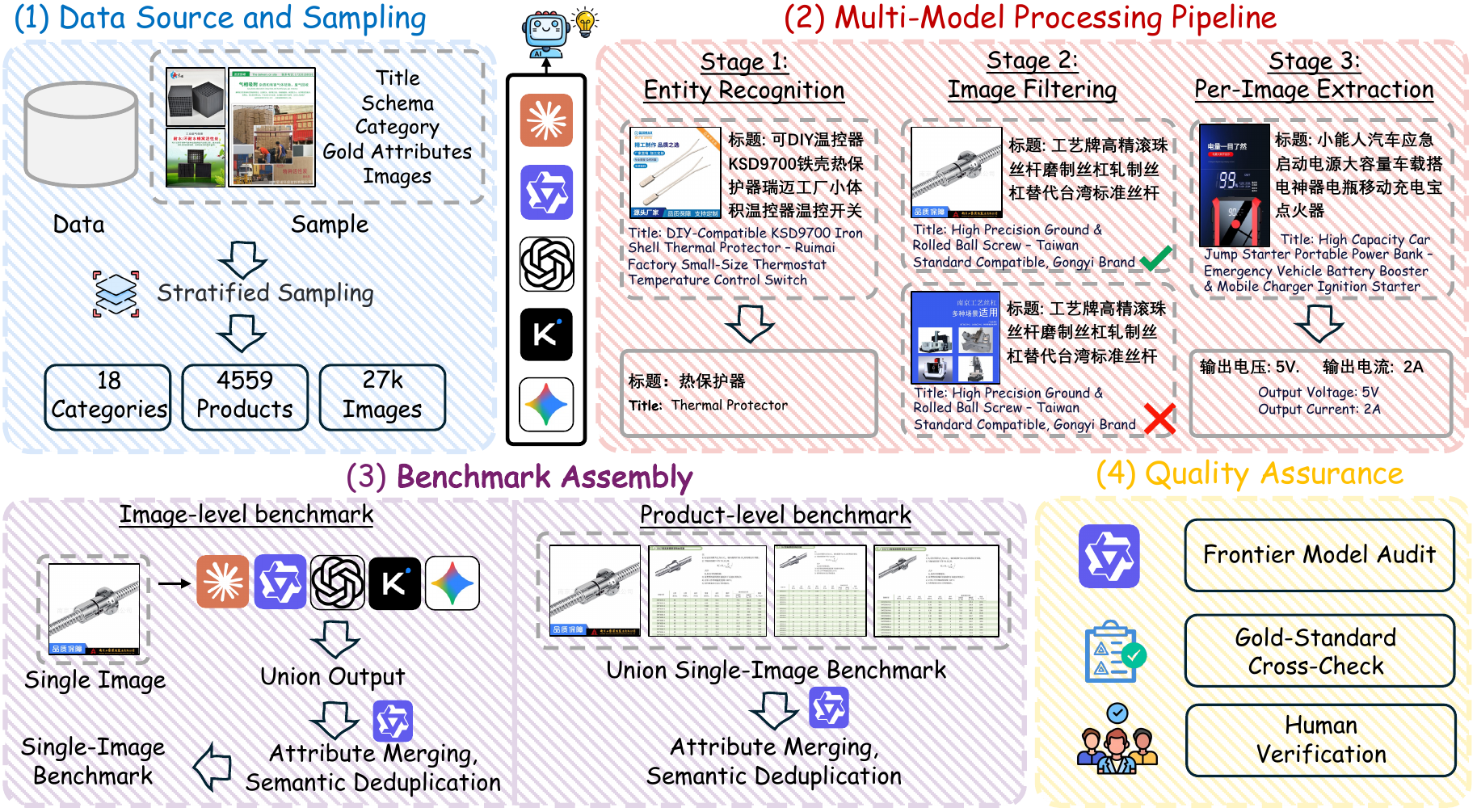}
\caption{Overview of the \bench{} Construction Pipeline. Product profiles are collected via stratified sampling (\S\ref{sec:sampling}). Five MLLMs independently execute a three-stage annotation pipeline (\S\ref{sec:annotation}). Results are assembled into image-level and product-level benchmarks via union and semantic deduplication (\S\ref{sec:assembly}), followed by three-tier quality assurance (\S\ref{sec:quality}).}
\label{fig:pipeline}
\end{figure}

% ============================================================
% 2. RELATED WORK
% ============================================================
\section{Related Work}
\label{sec:related}

\paragraph{Multimodal LLM Benchmarks.}
Most MLLM benchmarks frame evaluation as per-question correctness: MMMU \citep{yue2024mmmu} and MMMU-Pro \citep{yue2024mmmupro} measure college-level multimodal reasoning, MMBench \citep{liu2023mmbench} spans 20 ability dimensions, and OCRBench \citep{liu2024ocrbench} targets text recognition accuracy. Multi-image benchmarks such as MMIU \citep{meng2025mmiu} and MMRB \citep{cheng2025mmrb} extend evaluation to cross-image reasoning across diverse tasks. A complementary line of work probes whether models genuinely ground their answers in visual context rather than language priors, e.g., resolving textual ambiguity from paired images \citep{wang2026mma,pan2026vida,tang2026evos}. However, these benchmarks assess general visual understanding through closed-form questions, whereas industrial product understanding demands structured extraction from domain-specific, multi-image inputs---a fundamentally different evaluation paradigm.

\paragraph{Attribute Value Extraction.}
AVE has evolved from sequence labeling on product titles \citep{zheng2018opentag,xu2019scaling} through multi-source text datasets \citep{yang2022mave} to multimodal approaches integrating visual features \citep{zhu2020multimodal}. Recent work has expanded along several axes: ImplicitAVE \citep{wang2024implicitave} benchmarks implicit extraction via generative MLLMs, EIVEN \citep{eiven2024} and MICE \citep{mice2025} improve multimodal extraction through contrastive learning and captioning augmentation, and VideoAVE \citep{cheng2025videoave} extends extraction to product videos. However, existing AVE work does not address the industrial domain, where extraction requires interpreting domain-encoded values (e.g., ``DN20'' for nominal diameter), handling product-specific attribute schemas, and integrating specifications dispersed across multiple heterogeneous images.

\paragraph{Industrial AI Benchmarks.}
Existing industrial AI benchmarks focus on visual perception or text-only knowledge. MVTec AD \citep{bergmann2019mvtec} evaluates surface anomaly detection; FORGE \citep{forge2026} benchmarks manufacturing tasks such as material sorting and assembly recognition. These assess whether models can \textit{see} defects and distinguish parts, but not whether they can \textit{understand} product specifications. On the text side, IndustryBench \citep{industrybench2026} probes industrial domain knowledge through text-only QA without visual input. In the adjacent e-commerce setting, EcomBench \citep{min2025ecombench} evaluates foundation agents on realistic shopping tasks, but targets agentic decision-making rather than multimodal specification understanding.

In summary, existing benchmarks address visual reasoning, attribute extraction, and industrial knowledge in isolation, but none targets the intersection where all three converge: extracting structured technical attributes from multi-image industrial product profiles.

% ============================================================
% 3. TASK DEFINITION
% ============================================================
\section{Task Definition}
\label{sec:task}

To enable systematic evaluation of the challenges identified above, we formalize \textit{multimodal industrial product attribute extraction} as follows.

\paragraph{Input.}
An industrial product profile $x = (t, e, c_1, c_l, \{I^{(1)}, \ldots, I^{(n)}\}, S_x)$, consisting of a product title $t$, an entity description $e$ identifying the target product, top-level category $c_1$, leaf category $c_l$, a set of product images $\{I^{(i)}\}_{i=1}^{n}$, and a product-specific attribute schema $S_x$ defining permissible property names.

\paragraph{Output.}
A set of property-value pairs $\mathcal{Y} = \{(p_j, v_j)\}$, where $p_j \in S_x$ is a property name and $v_j$ is the corresponding value extracted from the visual and textual input.

We define two evaluation settings of increasing difficulty:

\paragraph{Single-Image Extraction.}
Given a single product image along with the product metadata and attribute schema, the model extracts all property-value pairs visible in that image. This setting isolates the model's core capabilities in text recognition, visual reasoning, and domain knowledge application.

\paragraph{Multi-Image Extraction.}
Given the complete product profile with all associated images, the model produces a unified set of property-value pairs for the entire product. This setting additionally tests the model's ability to locate relevant information across multiple images, filter irrelevant content, and integrate attributes from different sources.

% ============================================================
% 4. MIP-BENCH
% ============================================================
\section{\bench{}}
\label{sec:benchmark}

\subsection{Overview}
\label{sec:overview}

To make completeness measurable, the benchmark must define a target record that exceeds what any single model can produce---otherwise evaluation only measures agreement with one model's ceiling rather than true specification coverage. \bench{} achieves this through multi-model consensus: five MLLMs independently annotate the same products and their outputs are unioned, producing a recall ceiling substantially above any individual annotator.

\bench{} is sourced from a large-scale Chinese industrial e-commerce platform covering 18 top-level categories. Table~\ref{tab:dataset_stats} summarizes the key statistics. The benchmark comprises \textbf{4{,}559 products}, \textbf{27{,}652 valid images}, and \textbf{103{,}703 product-level annotations} across 3{,}564 unique property names, supporting both image-level (182{,}527 annotations) and product-level evaluation. All product profiles, schemas, and annotations are in Chinese; evaluated models extract Chinese-language property-value pairs.

\begin{table}[t]
\centering
\tighttable
\setlength{\tabcolsep}{6pt}
\begin{tabular}{@{}lr@{}}
\toprule
\textbf{Statistic} & \textbf{Value} \\
\midrule
Products & 4{,}559 \\
Valid images & 27{,}652 \\
Top-level categories & 18 \\
Unique property names & 3{,}564 \\
Image-level annotations & 182{,}527 \\
Product-level annotations & 103{,}703 \\
Median images per product & 5 (IQR: 2--8) \\
Median attrs per product & 15 (IQR: 7--30) \\
\bottomrule
\end{tabular}
\caption{Key Statistics of \bench{}.}
\label{tab:dataset_stats}
\end{table}

\subsection{Construction Pipeline}
\label{sec:construction}

Constructing a reliable benchmark for this task is non-trivial due to three interrelated challenges: (1)~\textbf{data scarcity}---publicly available multi-image industrial product profiles with structured attribute annotations are difficult to obtain; (2)~\textbf{schema complexity}---industrial products require product-specific attribute schemas (e.g., pressure rating for valves vs.\ rated current for circuit breakers); and (3)~\textbf{annotation difficulty}---many attributes are implicit (e.g., ``DN20'' denotes nominal diameter, ``304'' a steel grade), making purely manual annotation at scale prohibitively expensive. We address these challenges with a semi-automated pipeline that combines multi-model consensus with systematic quality assurance, as illustrated in Figure~\ref{fig:pipeline}.

\subsubsection{Data Source and Sampling}
\label{sec:sampling}

We source product profiles from a major Chinese industrial e-commerce platform, covering 18 top-level categories including hardware and tools, chemicals, electrical equipment, machinery, safety and protection, and packaging. Each profile consists of a product title, top-level and leaf-level category names, multiple product images, a product-specific attribute schema, and a set of verified but non-exhaustive standard attributes.

To ensure category diversity, we employ \textbf{stratified sampling}: for each leaf-level product category, we randomly sample a fixed number of products using a fixed seed. This yields a balanced dataset that avoids the long-tail dominance of popular categories, ultimately producing 4{,}559 products with 27{,}652 images.

\subsubsection{Multi-Model Annotation Pipeline}
\label{sec:annotation}

We run the entire annotation pipeline independently with $K=5$ MLLMs (GPT-5.4, Claude Opus 4.6, Gemini 3.1 Pro, Kimi-K2.5, and Qwen 3.5 Plus). Each model receives the product title, category names, product images, and product-specific attribute schema, and executes a three-stage pipeline that mirrors how a domain expert reviews a product profile: first identify what the product is, then determine which images are informative, then extract specifications from each valid image.

\paragraph{Stage 1: Entity Recognition.}
Given the product title $t$ and product images, the model identifies the core product entity $e$ (e.g., ``stainless steel two-piece ball valve''). This anchoring prevents \textit{entity drift}---extracting attributes from complementary products, installation contexts, or product-line overviews that frequently appear alongside the target product.

\paragraph{Stage 2: Image Filtering.}
For each image $I^{(i)}$, the model determines whether it contains attribute-relevant information for entity $e$, marking images such as factory facilities, marketing banners, and unrelated products as invalid. This pre-extraction filtering prevents \textit{hallucination cascades} by eliminating a primary source of false annotations. Approximately 31\% of candidate images are filtered out.

\paragraph{Stage 3: Per-Image Attribute Extraction.}
Each valid image is processed by an extraction agent that outputs structured property-value pairs, conditioned on entity $e$ and the product-specific attribute schema $S_x$. Each extracted attribute includes a property name, value, and extraction rationale. When multiple valid values exist for the same property, each value is represented as a separate property-value pair.

\begin{table*}[t]
\centering
\tighttable
\setlength{\tabcolsep}{4.4pt}
\begin{tabular}{@{}lcrrrrrrrrr@{}}
\toprule
& & \multicolumn{3}{c}{\textbf{Multi-Image}} & \multicolumn{3}{c}{\textbf{Single-Image}} & \multicolumn{3}{c}{\textbf{$\Delta$ (Multi $-$ Single)}} \\
\cmidrule(lr){3-5} \cmidrule(lr){6-8} \cmidrule(lr){9-11}
\textbf{Model} & \textbf{Rank} & Prec. & Rec. & F1 & Prec. & Rec. & F1 & Prec. & Rec. & F1 \\
\midrule
\multicolumn{11}{@{}l}{\textit{Closed-source}} \\
\addlinespace[2pt]
\rowcolor{bestrow}
Gemini 3.1 Pro & (1) & \textbf{93.8} & \textbf{49.9} & \textbf{65.1} & \textbf{94.0} & 65.4 & \underline{77.1} & $-$0.2 & $-$15.5 & $-$12.0 \\
GPT-5.4 & (3) & 86.3 & 46.6 & 60.5 & 82.7 & 55.3 & 66.2 & +3.6 & $-$8.7 & $-$5.7 \\
Qwen 3.5 Plus & (4) & 88.1 & 45.4 & 59.9 & 82.9 & \textbf{79.7} & \textbf{81.3} & +5.2 & $-$34.3 & $-$21.4 \\
Claude Opus 4.6 & (5) & 88.2 & 42.3 & 57.2 & \underline{85.1} & 58.6 & 69.4 & +3.1 & $-$16.3 & $-$12.2 \\
\midrule
\multicolumn{11}{@{}l}{\textit{Open-source}} \\
\addlinespace[2pt]
\rowcolor{bestrow}
Qwen 3.5-397B-A17B (MoE) & (2) & 88.2 & \underline{48.6} & \underline{62.7} & 80.6 & \underline{72.0} & 76.0 & +7.6 & $-$23.4 & $-$13.3 \\
Kimi-K2.5-1T-A32B (MoE) & (6) & 88.6 & 41.7 & 56.7 & 79.0 & 64.3 & 70.9 & +9.6 & $-$22.6 & $-$14.2 \\
Qwen 3.5-27B (Dense) & (7) & 88.0 & 40.8 & 55.8 & 78.8 & 65.4 & 71.5 & +9.2 & $-$24.6 & $-$15.7 \\
Qwen 3.5-122B-A10B (MoE) & (8) & \underline{88.8} & 34.9 & 50.1 & 77.6 & 64.6 & 70.5 & +11.2 & $-$29.7 & $-$20.4 \\
Qwen 3.5-35B-A3B (MoE) & (9) & 86.0 & 11.7 & 20.6 & 75.1 & 63.3 & 68.7 & +10.9 & $-$51.6 & $-$48.1 \\
\bottomrule
\end{tabular}
\caption{Main Results (\%) on \bench{}. Rank is determined by multi-image F1. Single-image precision, recall, and F1 are included as secondary references for localized extraction ability; $\Delta$ columns report multi-image minus single-image performance. Bold and underlined values mark the best and second-best metric scores, respectively.}
\label{tab:main_results}
\end{table*}

\subsubsection{Benchmark Assembly}
\label{sec:assembly}

We assemble the benchmark at two granularities from the five independent model runs.

\paragraph{Image-level benchmark.}
\label{sec:single_bench}
For each image, we pool all property-value pairs proposed by the five models. We take the union rather than majority vote because the goal is to maximize the recall ceiling of the benchmark: a valid attribute found by even one model should appear in the target record, otherwise the benchmark can only measure agreement with the majority rather than true specification coverage. We then normalize surface forms to merge formatting variants (e.g., ``304 stainless steel'' and ``stainless steel 304'') and use Qwen 3.6 Plus to semantically deduplicate near-equivalent entries.

\paragraph{Product-level benchmark.}
\label{sec:multi_bench}
To construct product-level annotations for each individual product, we merge the single-image benchmark attributes across all images of that product. When the same property appears with different values across images, we retain all values rather than forcing a single resolution, reflecting that industrial products may have multiple valid specification values (e.g., multiple models or supported voltage ratings). The aggregated set undergoes a second round of semantic deduplication to remove near-duplicates introduced during cross-image merging.

\subsubsection{Quality Assurance}
\label{sec:quality}

We employ a three-tier quality assurance mechanism to ensure annotation reliability. Each tier targets a different class of errors, and the tiers are applied sequentially so that cheaper automated checks handle the bulk of filtering before expensive human review.

\paragraph{Tier 1: Frontier Model Audit.}
An independent frontier model (Qwen 3.6 Plus, which is not among the nine evaluated models) reviews all candidate annotations. The audit applies four rule categories: (1)~hallucinated attributes with no visual evidence, (2)~entity misattribution where the attribute belongs to a different product in the image, (3)~property name--value type mismatches (e.g., a color field containing a part number), and (4)~semantic implausibility. Annotations flagged by any rule are marked as dropped and excluded from the benchmark. This tier filters 57{,}478 out of 240{,}054 candidate annotations (23.9\%).

\paragraph{Tier 2: Gold-Standard Cross-Check.}
Each product is associated with a verified but non-exhaustive set of standard attributes. For properties covered by these gold attributes, we directly replace the corresponding LLM-generated annotations with the verified values.

\paragraph{Tier 3: Human Verification.}
Domain experts perform spot-check verification on a random 10\% sample of annotated products, checking both attribute correctness and coverage of important specifications. The verification pass rate is 96.7\%, confirming the effectiveness of the preceding automated tiers.

Because the five annotator models also appear among the evaluated models, we analyze annotation overlap and discuss potential self-evaluation bias in Appendix~\ref{app:annotation_overlap}.

% ============================================================
% EVALUATION PROTOCOL
% ============================================================
\section{Evaluation Protocol}
\label{sec:eval_protocol}

\paragraph{Matching Strategy.}
We evaluate predicted property-value pairs against benchmark pairs within the same image or product scope. Property names require exact string match. Values are matched with a cascaded strategy: we first apply rule-based normalization (Unicode NFKC, case folding, punctuation/space removal, unit normalization, and numeric canonicalization) and accept exact normalized matches or character-level F1 above 0.6; remaining candidates are judged by Qwen 3.6 Plus for semantic equivalence. Predictions whose property names are absent from the benchmark are counted as extra-property errors.

\paragraph{Metrics.}
We report precision, recall, and F1 at both image and product levels (formal definitions in Appendix~\ref{app:metrics}). Precision is the fraction of predictions that are matched; recall is the fraction of benchmark pairs matched by at least one prediction, with all instances in the evaluation scope included in the denominator. We additionally report error-type breakdowns in Appendix~\ref{app:error_analysis}. Further details on property name and value matching are provided in Appendix~\ref{app:eval_details}.

% ============================================================
% 6. EXPERIMENTS
% ============================================================
% ============================================================
% EXPERIMENTS (standalone file — \input{experiments} from main.tex)
% ============================================================
\section{Experiments}
\label{sec:experiments}

\begin{figure}[t]
\centering
\includegraphics[width=0.72\textwidth]{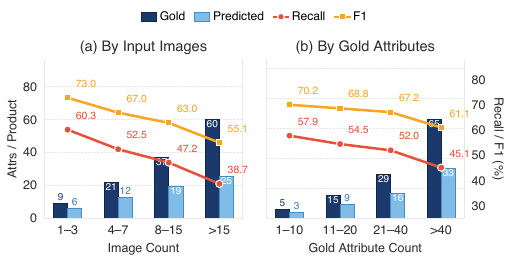}
\caption{Bottleneck Analysis for Gemini 3.1 Pro. Bars show the average number of gold and predicted property-value pairs per product, while lines show recall and F1. As the number of input images and benchmark attributes increases, gold evidence grows faster than model outputs, and recall/F1 decline.}
\label{fig:bottleneck_gemini}
\end{figure}

\subsection{Experimental Setup}

\paragraph{Models.}
We evaluate nine MLLMs, including four closed-source models (Gemini 3.1 Pro, Qwen 3.5 Plus, GPT-5.4, Claude Opus 4.6) and five open-source models (Kimi-K2.5-1T-A32B and four Qwen 3.5 variants: 27B, 35B-A3B, 122B-A10B, and 397B-A17B).

\paragraph{Settings.}
Our primary setting is \textbf{multi-image extraction}, where a model receives all valid images of a product and outputs product-level property-value pairs. We also report \textbf{single-image extraction}, where images are processed independently, as a diagnostic reference for localized extraction ability. Unless otherwise specified, we use the full extraction prompt and enable thinking mode for models that support it.

% ============================================================
\subsection{Main Results}
\label{sec:main_results}

Table~\ref{tab:main_results} presents the main results.

On the multi-image task, Gemini 3.1 Pro leads (F1 = 65.1\%), followed by Qwen 3.5-397B-A17B (62.7\%) and GPT-5.4 (60.5\%). The dominant pattern across all models is a precision--recall asymmetry: precision is consistently high (86--94\%), but even the best model recovers only half of the product-level benchmark attributes. The ranking is therefore recall-driven, with recall ranging from 11.7\% to 49.9\% while precision varies within an 8-point band. This pattern holds across all 18 categories (Appendix~\ref{app:category}).

Qwen 3.5 Plus sharpens the distinction between localized extraction and product-level understanding: it achieves the best single-image F1 (81.3\%) but drops to fourth on the multi-image task. The $\Delta$ columns quantify this gap systematically. For most models, moving from single-image to multi-image extraction costs 15--34 percentage points of recall; Qwen 3.5-35B suffers a near-complete collapse (63.3\% $\to$ 11.7\%). Crucially, precision holds or rises---the cost is paid entirely in recall. Models do not become less accurate with more images; they become less \textit{complete}, missing specifications in dense tables or distributed across less salient images.

The remaining subsections diagnose the sources of this completeness gap.

% ============================================================
\subsection{Bottlenecks: Image Count and Specification Density}
\label{sec:bottleneck}

Where does the recall gap originate? We stratify multi-image results along two axes: the number of input images (cross-image localization difficulty) and the number of benchmark attributes per product (specification density). Figure~\ref{fig:bottleneck_gemini} reports stratified results for Gemini 3.1 Pro; the same trend holds across models.

Both factors substantially affect recall. When the number of input images increases from 1--3 to more than 15, Gemini's recall drops from 60.3\% to 38.7\% and F1 from 73.0\% to 55.1\%. The model does produce more attributes for image-heavy products, but not proportionally: products in the largest image bucket contain 60.1 gold attributes on average, yet the model outputs only 25.6 predictions. Cross-image evidence localization is a direct source of under-extraction.

Attribute density shows the same pattern. For products with more than 40 gold attributes, Gemini outputs 33.0 predictions for 65.9 gold attributes, recovering only 45.1\%. In both cases, precision remains stable while recall declines---models stop extracting before they have exhausted the available evidence.

% ============================================================
\subsection{Semantic Subtype Analysis of Industrial Attributes}
\label{sec:subtype}

The bottleneck analysis above shows \textit{how much} models miss as complexity grows; we now ask \textit{what kinds} of attributes are most affected. We classify the 753 distinct property types into four semantic subtypes based on the cognitive demand they place on the extractor, defined independently of observed model difficulty: \textit{direct standardized} (single-value fields readable via OCR or fixed formats), \textit{domain knowledge} (interpretation-heavy industrial terminology, standards, codes, and material grades), \textit{multi-value compound specifications} (ranges, lists, and composite dimensions), and \textit{visual reasoning} (structural cues, appearance-based judgments, and diagram interpretation).

\begin{table}[t]
\centering
\tighttable
\setlength{\tabcolsep}{5pt}
\begin{tabular}{@{}lrrrrr@{}}
\toprule
\textbf{Subtype} & \textbf{\#Attr} & \textbf{\#Bench} & \textbf{Prec.} & \textbf{Rec.} & \textbf{F1} \\
\midrule
Direct std. & 346 & 40{,}855 & \textbf{90.7} & \textbf{45.9} & \textbf{60.9} \\
Domain know. & 182 & 11{,}451 & 88.1 & 44.1 & 58.8 \\
Multi-value & 124 & 27{,}245 & 89.1 & 42.8 & 57.9 \\
Visual reason. & 101 & 5{,}857 & 85.9 & 36.6 & 51.3 \\
\bottomrule
\end{tabular}
\caption{Semantic Subtype Results (\%) in the Main Multi-Image Setting. Metrics are pooled over six representative multi-image model outputs. Subtypes are assigned from property semantics, independent of empirical model difficulty.}
\label{tab:semantic_subtype}
\end{table}

Table~\ref{tab:semantic_subtype} reveals a recall gradient that tracks cognitive demand. Direct standardized fields are easiest yet still reach only 45.9\% recall, confirming that the completeness bottleneck extends beyond complex reasoning. Recall degrades further for domain knowledge (44.1\%) and multi-value compound specifications (42.8\%), and is lowest for visual reasoning (36.6\%)---a 9.3-point spread that shows technical interpretation and dense-list aggregation compound the extraction gap. Per-model breakdowns (Appendix~\ref{app:subtype_per_model}) confirm this ordering is consistent across models, with the gap between stronger and weaker models widening on reasoning-intensive subtypes.

% ============================================================
\subsection{Scaling Behavior}
\label{sec:scaling}

The Qwen 3.5 family---five models sharing the same architecture and training methodology---enables a controlled study of how scale affects the single-image/multi-image gap.

\begin{table}[t]
\centering
\tighttable
\setlength{\tabcolsep}{5pt}
\begin{tabular}{@{}l@{\hskip 8pt}r@{\hskip 12pt}rr@{\hskip 12pt}rr@{}}
\toprule
& & \multicolumn{2}{c}{\textbf{Multi-Image}} & \multicolumn{2}{c}{\textbf{Single-Image}} \\
\cmidrule(lr){3-4} \cmidrule(lr){5-6}
\textbf{Model} & \textbf{Active} & Rec. & F1 & Rec. & F1 \\
\midrule
Qwen 3.5 Plus & -- & 45.4 & 59.9 & \textbf{79.7} & \textbf{81.3} \\
397B-A17B & 17B & \textbf{48.6} & \textbf{62.7} & 72.0 & 76.0 \\
27B (Dense) & 27B & 40.8 & 55.8 & 65.4 & 71.5 \\
122B-A10B & 10B & 34.9 & 50.1 & 64.6 & 70.5 \\
35B-A3B & 3B & 11.7 & 20.6 & 63.3 & 68.7 \\
\bottomrule
\end{tabular}
\caption{Scaling Analysis (\%) Within the Qwen 3.5 Family. Multi-image extraction exhibits a much sharper scale sensitivity than single-image extraction.}
\label{tab:scaling}
\end{table}

Table~\ref{tab:scaling} reveals two distinct regimes. In the single-image setting, performance improves smoothly with scale: F1 rises from 68.7\% (35B-A3B) to 76.0\% (397B-A17B), with Qwen 3.5 Plus reaching 81.3\%. Notably, the 27B dense model outperforms the 122B-A10B MoE variant, indicating that active parameter count matters more than total parameters for localized extraction.

The multi-image setting reveals a far sharper scale dependence. The smallest model (35B-A3B, 3B active) nearly collapses to 20.6\% F1, while 397B-A17B reaches 62.7\%---a 42-point spread compared to only 7.3 points in single-image F1. The sharpest jump occurs between 10B and 17B active parameters, where multi-image F1 gains 12.6 points versus 5.5 in single-image. Cross-image integration---maintaining attention, locating distributed evidence, and assembling complete specifications---is substantially more scale-sensitive than localized extraction.

\begin{figure}[t]
\centering
\includegraphics[width=0.92\textwidth]{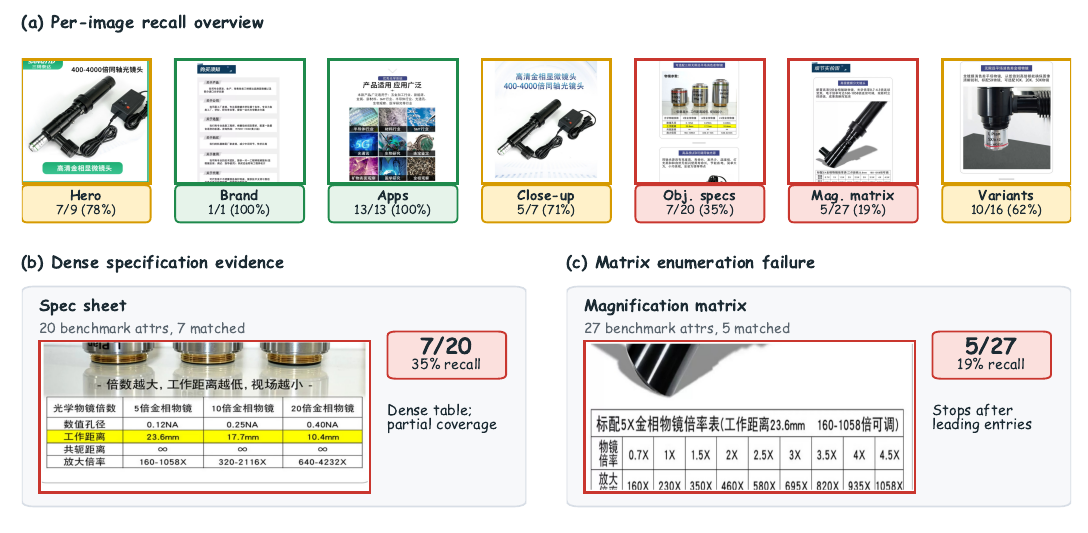}
\caption{Case Product Evidence Layout. Panel (a) shows per-image recall for the product's 7 valid images, with labels reporting matched / benchmark attributes and recall percentage. Panels (b) and (c) zoom into the two spec-dense images where recall collapses: the objective-spec sheet and the magnification matrix. Green / amber / red denote $\ge$85\%, 50--85\%, and $<$50\% recall, respectively.}
\label{fig:case_study}
\end{figure}

% ============================================================
\subsection{Ablations: Prompting and Inference Mode}
\label{sec:ablations}

We conduct two ablations with Qwen 3.5 Plus to examine whether inference and prompting choices change the precision--recall tradeoff. These ablations do not change the benchmark or evaluation protocol.

\begin{table}[t]
\centering
\tighttable
\setlength{\tabcolsep}{5pt}
\begin{tabular}{@{}llrrrr@{}}
\toprule
\textbf{Ablation} & \textbf{Setting} & \textbf{Prec.} & \textbf{Rec.} & \textbf{F1} & \textbf{\#Pred.} \\
\midrule
\multirow{2}{*}{Inference} & Thinking & \textbf{88.1} & 45.4 & 59.9 & 50{,}334 \\
& No-think & 86.9 & 46.5 & 60.6 & 56{,}180 \\
\midrule
\multirow{2}{*}{Prompt} & Full & \textbf{88.1} & 45.4 & 59.9 & 50{,}334 \\
& Simple & 77.3 & \textbf{52.6} & \textbf{62.6} & 73{,}777 \\
\bottomrule
\end{tabular}
\caption{Ablation Results for Qwen 3.5 Plus on Multi-Image Extraction. No-thinking and simpler prompting both increase recall, but at different precision costs.}
\label{tab:ablation_results}
\end{table}

Table~\ref{tab:ablation_results} shows that both ablations shift the precision--recall tradeoff without eliminating the completeness bottleneck. Removing thinking slightly increases recall (+1.1 pp) with a modest precision drop. The simpler prompt has a stronger effect: recall rises from 45.4\% to 52.6\% and F1 from 59.9\% to 62.6\%, but precision falls from 88.1\% to 77.3\%. The simple prompt elicits 47\% more predictions (50K $\to$ 74K) by removing schema constraints that act as implicit extraction gates, trading precision for coverage. This shows that the recall ceiling is partly prompt-induced, but even the most permissive setting still misses nearly half the target record. Notably, the thinking effect reverses in the single-image setting (Appendix~\ref{app:thinking_single}).

% ============================================================
\FloatBarrier
\subsection{Case Study: Anatomy of the Completeness Gap}
\label{sec:case_study}

The preceding analyses show that the multi-image gap is primarily a recall problem. We now examine one representative product to identify what the missing attributes look like. The product is a coaxial-light microscope objective from the \textit{instruments} category, with 7 valid images and 69 benchmark attributes. Under the multi-image setting, the top-ranked model attains \textbf{100\%} precision but only \textbf{45.0\%} recall: every extracted value is correct, yet more than half of the visible specifications are omitted.

\paragraph{Per-image recall follows evidence layout.}
Figure~\ref{fig:case_study} separates the product images into two regimes. On compact, well-segmented evidence---the brand logo, the 9-cell application-icon grid, and the hero shot---the model recovers 21 of 23 benchmark attributes ($>$91\%). On spec-dense evidence, recall collapses: the objective-spec sheet contains 20 benchmark attributes but yields only 7 matches (35\%), while the magnification matrix---a 9-column table listing objective and total magnifications side by side---contains 27 attributes but yields only 5 matches (19\%). The failure is therefore not caused by the mere presence of multiple images; it appears when a single image requires exhaustive enumeration from a dense specification layout.

\begin{table}[H]
\centering
\tighttable
\setlength{\tabcolsep}{4pt}
\begin{tabular}{@{}l@{\hskip 6pt}rrrl@{}}
\toprule
\textbf{Property} & \textbf{\#GT} & \textbf{\#Pred} & \textbf{\#Match} & \textbf{Example missed values} \\
\midrule
Application                & 13 & 13 & \textbf{13} & --- \\
Brand                      &  2 &  2 & \textbf{2}  & --- \\
\midrule
Objective magnification    & 14 &  4 & 4  & 0.7X, 1X, 1.5X, 2X, 2.5X, \ldots \\
Total magnification        & 15 &  5 & 5  & 160X, 230X, 350X, 460X, 580X, \ldots \\
Lens type                  &  7 &  2 & 2  & plan apochromat, metallographic, beam-splitter \\
Model code                 &  4 &  0 & \textbf{0}  & L Plan 5X/0.12, 10X/0.25, 20X/0.40 \\
Color                      &  5 &  1 & 1  & red, yellow, green, silver \\
Optical structure          &  5 &  3 & 2  & coaxial-light, plan, apochromatic \\
\bottomrule
\end{tabular}
\caption{Selected Per-Property Breakdown for the Case Product. Single-instance fields (top block) are recovered exhaustively, whereas variant lists and product-matrix fields (bottom block) are systematically under-extracted from dense specification images.}
\label{tab:case_study_props}
\end{table}

\paragraph{Failures concentrate in enumerative fields.}
Table~\ref{tab:case_study_props} shows the same split at the property level. Single-instance fields are not the bottleneck: brand and application values are recovered exhaustively. The missing attributes instead concentrate in fields whose values form a list or product matrix. Among the 14 objective magnifications and 15 total magnifications in the matrix table, the model returns only the first 4--5 values and then stops, leaving the remaining visible entries unreported. The four manufacturer-code variants (L Plan 5X/0.12, L Plan 10X/0.25, L Plan 20X/0.40, WD10.4), shown in another specification image, are missed in their entirety.

\paragraph{Connection to the aggregate findings.}
This case instantiates the patterns in \S\ref{sec:bottleneck} and \S\ref{sec:subtype}. The model recognizes the relevant specification fields and reads several values correctly, but terminates before enumerating the complete set. The dominant error mode is thus premature stopping in multi-value extraction, rather than hallucination or broad OCR failure. This explains how high precision can coexist with low recall, and why the aggregate gap widens on dense, multi-value industrial specifications.

% ============================================================
% CONCLUSION
% ============================================================
\section{Conclusion}
\label{sec:conclusion}

We presented \bench{}, the first multi-image benchmark for evaluating MLLM understanding of industrial products through structured attribute extraction. Spanning 4{,}559 products, 27{,}652 images, and 103{,}703 annotations across 18 industrial categories, \bench{} reveals a consistent completeness gap: current MLLMs achieve high precision but recover at most half of the product-level specifications, with the deficit concentrating at the intersection of cross-image evidence integration, specification density, and domain-specific reasoning. Prompt and scale adjustments shift this tradeoff without closing it, suggesting that the bottleneck is not merely an artifact of a single prompt or inference mode. We release \bench{} to the community as a testbed for closing this gap---bridging multimodal reasoning, domain knowledge, and multi-image comprehension in real-world industrial settings.

% ============================================================
% LIMITATIONS & ETHICS
% ============================================================
\phantomsection
\section*{\texorpdfstring{\textdagger\ Author Contributions}{Author Contributions}}
\label{sec:authors}
\paragraph{Project Leader:} Liang Ding.
\paragraph{Core contributors:}
Haonan Qi, Jin Cao, Yongqi Zhang, Xintong Wang\footnotemark[1], Weidong Tang, Bin Chen, Chengfu Huo, Liang Ding\footnotemark[1].
\paragraph{Contributors:}
Haojun Pan, Hengyu You, Jing Li, Yingde Wang.

\footnotetext[1]{Corresponding to: \email{hanfeng.wxt@alibaba-inc.com}, and \email{zuorui.dl@alibaba-inc.com}}

\section*{Limitations}

The dataset contains only Chinese-language product profiles, which may limit generalizability to other languages. The LLM-in-the-loop annotation may inherit systematic biases from the MLLMs used for candidate generation, though we mitigate this through multi-model consensus and human verification. Our main results use a single prompt per evaluation setting; the ablation in \S\ref{sec:ablations} shows that prompt choice can shift F1 by up to 2.7 percentage points, so model rankings may be partially prompt-sensitive.

\section*{Ethics Statement}

\bench{} is constructed from industrial product profiles rather than personal or conversational data. We do not intentionally collect private user information; the images and text describe commercial products, specifications, labels, and packaging. Because product images may contain brand names, logos, and supplier-provided markings, the released benchmark should be used for research on multimodal product understanding rather than for misrepresentation, counterfeit detection claims, or automated procurement decisions without human oversight. The benchmark annotations are produced with MLLM assistance and may still contain residual errors despite multi-model consensus, frontier-model audit, gold-standard cross-checking, and human verification. AI tools were also used to assist paper writing and editing, with all content reviewed by the authors. We therefore encourage users to treat model outputs as decision-support signals, especially in safety-critical industrial settings where incorrect specifications could have operational consequences.

\bibliographystyle{plainnat}
\bibliography{custom}

\appendix

\section{Formal Metric Definitions}
\label{app:metrics}

This section provides formal definitions for the metrics described in \S\ref{sec:eval_protocol}.

Let $\mathcal{G}$ be the set of gold CPV pairs and $\mathcal{P}$ the set of predicted CPV pairs for a given evaluation scope. Let $\mathcal{M}_G \subseteq \mathcal{G}$ and $\mathcal{M}_P \subseteq \mathcal{P}$ be the matched subsets found by the greedy matching algorithm. Then:
\begin{align}
\text{Precision} &= \frac{|\mathcal{M}_P|}{|\mathcal{P}|}, \quad
\text{Recall} = \frac{|\mathcal{M}_G|}{|\mathcal{G}|}. \\
\text{F1} &= \frac{2 \cdot \text{Precision} \cdot \text{Recall}}{\text{Precision} + \text{Recall}}.
\end{align}

The matching algorithm iterates over predicted pairs in sorted order and greedily assigns each to the first unmatched gold pair with the same property name and a soft-matching value. Exact value matches are preferred over soft matches when multiple candidates exist.

\section{Additional Results}
\label{app:additional_results}

\subsection{Error Patterns}
\label{app:error_analysis}

The main results (\S\ref{sec:main_results}) show that errors are dominated by omissions rather than incorrect values. Here we decompose the non-omission errors into two types: \textit{extra property} errors, where the model predicts a property name absent from the benchmark, and \textit{value mismatch} errors, where the property name is correct but the predicted value does not match.

\begin{table}[htbp]
\centering
\tighttable
\setlength{\tabcolsep}{5pt}
\begin{tabular}{@{}lrrr@{}}
\toprule
\textbf{Model} & \textbf{Extra} & \textbf{ValMis.} & \textbf{Rule} \\
\midrule
Gemini 3.1 Pro & 4.3 & 1.7 & \textbf{92.3} \\
Qwen 3.5 Plus & 9.8 & 6.8 & 85.1 \\
Claude Opus 4.6 & 11.7 & 3.2 & 82.8 \\
GPT-5.4 & 10.7 & 6.6 & 80.4 \\
Kimi-K2.5 & 15.3 & 5.7 & 75.7 \\
\bottomrule
\end{tabular}
\caption{Error Rates (\%) in Single-Image Extraction. \textit{Extra}: predictions with property name absent from benchmark; \textit{ValMis.}: value mismatch; \textit{Rule}: predictions matched by rule-based matching without LLM semantic judgment.}
\label{tab:app_error_analysis}
\end{table}

Value mismatch rates are comparatively small across models. This reinforces the completeness gap identified in the main results: the dominant error mode is missing attributes, not incorrect values.

\subsection{Category-Level Variation}
\label{app:category}

Table~\ref{tab:app_category_f1} extends the main multi-image results (\S\ref{sec:main_results}) by reporting Gemini 3.1 Pro's performance across all 18 top-level categories (12 shown; 6 categories with fewer than 300 benchmark annotations are omitted for space).

\begin{table}[htbp]
\centering
\tighttable
\setlength{\tabcolsep}{5pt}
\begin{tabular}{@{}l@{\hskip 10pt}rrrr@{}}
\toprule
\textbf{Category} & \textbf{Prec.} & \textbf{Rec.} & \textbf{F1} & \textbf{\#Bench} \\
\midrule
Rubber \& Plastics & 95.6 & 56.1 & 70.7 & 7{,}925 \\
Steel & 80.4 & 63.0 & 70.6 & 378 \\
Electronic Components & 93.3 & 52.7 & 67.4 & 7{,}053 \\
Instruments & 92.3 & 52.5 & 66.9 & 2{,}579 \\
Chemicals & 93.9 & 51.5 & 66.5 & 14{,}095 \\
Construction Materials & 95.7 & 50.2 & 65.8 & 2{,}168 \\
Hardware \& Tools & 93.5 & 49.5 & 64.7 & 26{,}782 \\
Electrical Equipment & 93.3 & 49.3 & 64.5 & 10{,}112 \\
Machinery & 92.7 & 48.7 & 63.8 & 14{,}543 \\
Safety \& Protection & 95.8 & 47.6 & 63.6 & 5{,}904 \\
Packaging & 95.0 & 42.7 & 58.9 & 7{,}029 \\
Textiles \& Leather & 95.0 & 35.3 & 51.4 & 3{,}677 \\
\bottomrule
\end{tabular}
\caption{Multi-Image Performance (\%) by Category for Gemini 3.1 Pro. \#Bench = benchmark annotations per category.}
\label{tab:app_category_f1}
\end{table}

Precision remains uniformly high across all categories (80--96\%), while recall varies substantially---from 56.1\% on Rubber \& Plastics to 35.3\% on Textiles \& Leather. The precision--recall asymmetry observed in the main results holds within every category: inter-category difficulty is driven entirely by variation in completeness, not accuracy.

\subsection{Per-Model Semantic Subtype Results}
\label{app:subtype_per_model}

Table~\ref{tab:app_subtype_per_model} disaggregates the pooled subtype results from \S\ref{sec:subtype} into individual models. The recall gradient that tracks cognitive demand in the main text is consistent across all six models: visual reasoning $<$ domain knowledge $\approx$ multi-value $<$ direct standardized. Notably, the inter-model recall spread varies by subtype: it is widest for multi-value (14.7 pp) and domain knowledge (12.6 pp) but narrowest for visual reasoning (5.4 pp), suggesting that visual reasoning is a shared ceiling rather than a differentiator among current models.

\begin{table}[htbp]
\centering
\tighttable
\setlength{\tabcolsep}{5pt}
\begin{tabular}{@{}l@{\hskip 8pt}rrrrrr@{}}
\toprule
& \textbf{Gem.} & \textbf{GPT} & \textbf{397B} & \textbf{Opus} & \textbf{Kimi} & \textbf{27B} \\
\midrule
Direct std. & \textbf{50.6} & 47.3 & 46.8 & 43.6 & 44.3 & 42.7 \\
Domain know. & \textbf{51.2} & 45.2 & 47.4 & 39.9 & 42.2 & 38.6 \\
Multi-value & \textbf{51.9} & 46.3 & 43.6 & 40.0 & 37.2 & 38.2 \\
Visual reason. & 36.1 & \textbf{39.9} & 39.4 & 33.9 & 35.8 & 34.5 \\
\bottomrule
\end{tabular}
\caption{Recall (\%) by Semantic Subtype and Model in Multi-Image Extraction. Precision is uniformly high (82--95\%) across all cells and omitted for clarity.}
\label{tab:app_subtype_per_model}
\end{table}

\section{Evaluation Prompt Templates}
\label{app:prompts}

We provide the core prompt templates used in evaluation. The released code contains the full Chinese prompts with the same structure and output schema.

\paragraph{Multi-image extraction prompt.}
The model is instructed to act as an industrial product attribute extraction expert. Inputs include all product images, the product entity, top-level and leaf categories, and the product-specific schema. The model must extract only schema-listed property-value pairs supported by image evidence, including visible product features, labels, nameplates, specification tables, certification marks, and technical diagrams. It must ignore unrelated products, factory or marketing images, packaging containers when they are not the target product, and marketing language. Multiple values for the same property are output as separate pairs, duplicate values across images are removed, units are preserved, and the output must be a JSON object with a \texttt{cpv\_results} list.

\paragraph{Single-image extraction prompt.}
The single-image prompt follows the same evidence and schema constraints, but the source is restricted to one image. Product title, category, and entity information are used only to identify the target product and cannot serve as direct evidence for attribute values.

\paragraph{Simple prompt ablation.}
The simplified prompt keeps only the essential instruction: given all product images and the product-specific schema, extract all visible or inferable schema-listed property-value pairs for the target industrial product and return them as JSON. It removes most detailed evidence, entity-drift, confidence calibration, and filtering rules.

\paragraph{Semantic matching judge prompt.}
When rule-based value matching fails, the judge receives a property name, one predicted value, and candidate benchmark values for the same property. It must decide whether the prediction is semantically equivalent to any benchmark value, allowing formatting, ordering, abbreviation, and unit-surface variation when meaning is unchanged, and return a JSON decision with the matched benchmark value and a short reason.

\section{Thinking Mode in Single-Image Extraction}
\label{app:thinking_single}

Table~\ref{tab:thinking_single} reports the thinking vs.\ no-thinking comparison for Qwen 3.5 Plus in the single-image setting. In contrast to the multi-image setting, where disabling thinking yields only a marginal recall gain of +1.1 pp (\S\ref{sec:ablations}), single-image extraction benefits substantially from explicit reasoning.

\begin{table}[htbp]
\centering
\tighttable
\setlength{\tabcolsep}{5pt}
\begin{tabular}{@{}lrrrr@{}}
\toprule
\textbf{Setting} & \textbf{Prec.} & \textbf{Rec.} & \textbf{F1} & \textbf{\#Pred.} \\
\midrule
Thinking & \textbf{82.9} & \textbf{79.7} & \textbf{81.3} & 166{,}994 \\
No-thinking & 79.1 & 70.5 & 74.6 & 165{,}264 \\
\midrule
$\Delta$ (No-thinking $-$ Thinking) & $-$3.8 & $-$9.2 & $-$6.7 & \\
\bottomrule
\end{tabular}
\caption{Thinking vs.\ No-Thinking for Qwen 3.5 Plus on Single-Image Extraction. Thinking improves both precision and recall, with the recall gain (+9.2 pp) driving most of the F1 improvement.}
\label{tab:thinking_single}
\end{table}

Thinking mode improves recall by 9.2 pp in the single-image setting, indicating that explicit reasoning helps the model more exhaustively enumerate attributes from a single image. The asymmetry between settings suggests that the multi-image bottleneck lies in evidence \textit{localization} across images rather than in per-image \textit{enumeration}---a challenge that thinking alone does not address.

\section{Evaluation Protocol Details}
\label{app:eval_details}

\paragraph{Property name matching.}
Property names are matched by exact string equality after Unicode NFKC normalization. Because the benchmark schema provides canonical property names and both annotation and evaluation are schema-constrained, near-synonym ambiguity is rare: models must output property names from the provided schema. In our evaluation, over 99\% of correct predictions match on the first exact-match pass.

\paragraph{Value matching.}
When normalized exact match fails, remaining candidates are judged by Qwen 3.6 Plus for semantic equivalence, allowing formatting, ordering, abbreviation, and unit-surface variation when meaning is unchanged. The cascaded design ensures that the LLM judge handles only a small residual fraction of comparisons, limiting its potential error impact on overall metrics.

\section{Annotation Overlap and Self-Evaluation Analysis}
\label{app:annotation_overlap}

Because all five annotator models also serve as evaluated models, we analyze annotation consensus to assess potential self-evaluation bias.

\paragraph{Consensus distribution.}
Table~\ref{tab:consensus} reports the fraction of final product-level benchmark annotations that were independently proposed by $k$ of the 5 annotator models. Over half (53.2\%) of annotations are supported by 3 or more models, and 71.5\% by at least 2 models. While 28.5\% of annotations come from a single model, these have survived the three-tier QA process (including frontier model audit and gold-standard cross-check), indicating that they represent valid long-tail attributes that only one model successfully identified.

\begin{table}[htbp]
\centering
\tighttable
\setlength{\tabcolsep}{5pt}
\begin{tabular}{@{}lrrr@{}}
\toprule
\textbf{Consensus} & \textbf{\#Annot.} & \textbf{\%} & \textbf{Cumul.} \\
\midrule
5/5 models & 15{,}219 & 14.3 & 14.3 \\
4/5 models & 17{,}080 & 16.0 & 30.3 \\
3/5 models & 24{,}461 & 22.9 & 53.2 \\
2/5 models & 19{,}502 & 18.3 & 71.5 \\
1/5 models & 30{,}438 & 28.5 & 100.0 \\
\midrule
Total & 106{,}700 & & \\
\bottomrule
\end{tabular}
\caption{Consensus Distribution of Product-Level Benchmark Annotations. Each row shows how many annotations were independently proposed by exactly $k$ of the 5 annotator models (after QA filtering).}
\label{tab:consensus}
\end{table}

\paragraph{Why self-evaluation bias is limited.}
Three design choices mitigate the concern that annotator models are advantaged during evaluation:
(1)~The benchmark uses the \textit{union} of five models' outputs, so no single model's annotations dominate---an evaluated model's own contributions are diluted by the other four annotators.
(2)~The independent QA tier (Tier 1) removes 23.9\% of candidates, including model-specific hallucinations, using a model (Qwen 3.6 Plus) that is \textit{not} among the nine evaluated models.
(3)~Gold-standard cross-check (Tier 2) replaces LLM-generated values with verified ground truth for properties covered by the platform's standard attributes, further decoupling the benchmark from any individual annotator model's outputs.
Moreover, the evaluation is recall-dominated: a model's primary challenge is finding attributes it \textit{missed} during annotation (attributes contributed by other models), not confirming attributes it already proposed. The result that no model exceeds 50\% multi-image recall---despite contributing to the benchmark---empirically confirms that the benchmark is substantially harder than any single model's coverage.

\end{document}